\documentclass[11pt]{article}
\usepackage{geometry}
\usepackage{coling2020}
\usepackage{times}
\usepackage{url}
\usepackage{latexsym}
\usepackage{microtype}

\usepackage{amssymb}
\usepackage{tikz,pgfplots}
\usepackage{multirow}
\usepackage{booktabs}

\usepackage{subcaption}

\colingfinalcopy 

\pgfplotsset{compat=1.16}
\title{SST-BERT at SemEval-2020 Task 1: Semantic Shift Tracing by Clustering in BERT-based Embedding Spaces}

\author{Vani K\\ \And
  Sandra Mitrovi\'c \\ 
\hspace{4cm}IDSIA, Lugano, Switzerland \\ 
\hspace{4cm}{\tt \{vanik, sandra, alessandro, fabio.rinaldi\}@idsia.ch} \\\And
  Alessandro Antonucci \\\And
  Fabio Rinaldi }

\date{}

\begin{document}
\maketitle
\begin{abstract}
    Lexical semantic change detection (also known as \textit{semantic shift tracing}) is a task of identifying words that have changed their meaning over time. Unsupervised semantic shift tracing, focal point of SemEval2020, is particularly challenging. Given the unsupervised setup, in this work, we propose to identify clusters among different occurrences of each target word, considering these as representatives of different word meanings. As such, disagreements in obtained clusters naturally allow to quantify the level of semantic shift per each target word in four target languages.  To leverage this idea, clustering is performed on contextualized (BERT-based) embeddings of word occurrences.  
    The obtained results show that our approach performs well both measured separately (per language) and overall, where we surpass all provided SemEval baselines. 
\end{abstract}

\section{Problem Setup}
\blfootnote{
    %
    %
    %
    %
     
    
     \hspace{-0.65cm}  
     This work is licensed under a Creative Commons 
     Attribution 4.0 International License.
     License details:
     \url{http://creativecommons.org/licenses/by/4.0/}.
}
Consider two corpora $\mathcal{C}_1$ and $\mathcal{C}_2$ for a same language but associated with different time stamps  (say, respectively, $t_1$ and $t_2>t_1$). Let $\mathcal{W}$ be a set of \emph{target} words occurring in both corpora. Each target word $w\in\mathcal{W}$ might assume multiple meanings, to be called \emph{senses}, within the two corpora. A pool of experts annotated a representative amount of occurrences with their corresponding senses. The problem we consider is to characterize the semantic shift related to those senses from one corpus to the other without having access to the expert annotations. In particular, we address the two following two subtasks:
\begin{itemize}
\item \textbf{Subtask 1}: Decide, for each $w \in \mathcal{W}$, whether or not $w$ gained or lost at least a \emph{sense} between $t_1$ and $t_2$. This is a binary decision task. We will denote this subtask as \textbf{(S1)}.
\item \textbf{Subtask 2}: Define, for the elements of $\mathcal{W}$, a measure of their degree of lexical semantic change between $t_1$ and $t_2$ and sort these elements consequently. This is a ranking task. This subtask will be referred to as \textbf{(S2)}.
\end{itemize}
We\footnote{Our team name in SemEval2020 competition is NLP@IDSIA.} describe two different methods able to address both subtasks. As both methods require a preprocessing step based on transformers, let us start from this preliminary operation.

\section{Preprocessing}
For the proposed approach, we used embeddings derived from BERT (Bidirectional Encoder Representations from Transformers) \cite{devlin2018bert} model to represent the text information in the corpora. BERT uses attention mechanism to learn the contextual relations and reads the input bidirectionally. It is an encoder-only model (as the goal is to generate a language model) opposed to transformers (encoder-decoder model)\cite{vaswani2017attention}. BERT is trained on two main objectives, masked language model (MLM) and next sentence prediction (NSP). \\The corpus data is initially segmented at sentence-level and the BERT Word Piece tokenizer is applied on these sentences, to get the token-level representations. BERT-base model with twelve transformer layers is used and we derived the final embedding by concatenating the final four layers. If a single word gets split by the tokenizer, we take the average embedding value of the sub-tokens. Thus, for each target word we extract the embeddings from all the sentences associated with it from both corpora, $\mathcal{C}_1$ and $\mathcal{C}_2$. For corpora other than English language, we used multilingual BERT models of respective languages. We used the pre-trained model to generate the embeddings for all experiments, since the task is completely unsupervised in nature.

Let us assume, for each $w\in\mathcal{W}$, that $\mathcal{S}_j:=\{ s^{(w)}_{ij} \}_{i=1}^{n^{(w)}_j}$ denote the $n^{(w)}_j$ sentences in corpus $\mathcal{C}_j$ where target word $w$ occurred, for each  $j=1,2$. The transformer maps those sentences into corresponding vectors of a $d$-dimensional space. Let us perform a single transformation for both $\mathcal{S}_1$ and $\mathcal{S}_2$ simultaneously. Denote as $x^{(w)}_{ij}\in\mathbb{R}^d$ the vector associated with $s^{(w)}_{ij}$, for each $i=1,\ldots,n_{w,j}$ and $j=1,2$.  We regard the relative distances between these vectors as proxy indicators of their semantic similarity. This helps to cluster the similar senses together. 
While different distances, such as cosine, Manhattan and Euclidean could be used to measure BERT embedding similarities, we adopt the Euclidean distance given the  findings in \cite{hewitt2019structural,inui2019proceedings,reif2019visualizing} which demonstrate that the syntax tree distance between two words in BERT embedding space corresponds to the square of the Euclidean distance. In the next two sections, in order to address subtasks (S1) and (S2), we focus on the vectors associated with the same target word $w$ and use their relative Euclidean distances as indicators of possible semantic shifts from one corpus to the other.

Both methods will be based on clustering algorithms used to cluster the vectors associated with a given target word of a single corpus or of the union of the two. We adopt the classical $k$-means clustering algorithm, which forms the clusters by attempting to minimize the intra-cluster variance. So called \emph{silhouette} method is used for the selection of the optimal number of clusters $k$ and the initialization (i.e., the position of the centroids before starting the algorithm) \cite{rousseeuw1987silhouettes}. Accordingly, given a value of $k$, we compute in the corresponding cluster, the means of both the nearest-inter cluster distance and the nearest-cluster distance. The difference between these two quantities normalized by the maximum of the two is used as a fitness score to be maximized in order to select the optimal value of $k$. The same approach is used to determine the initial centroids. In this case, we use the optimal  $k$ value and run the  $k$-means algorithm for  $N$ iterations, to determine the best centroids. 

\section{Method 1: Joint Clustering Vectors of Both Corpora}
Let us focus on a particular target word $w \in \mathcal{W}$. Accordingly, for the sake of readability, denote its vectors in corpus $\mathcal{C}_j$ simply as $\mathcal{X}_j:=\{ x_{ij} \}_{i=1}^{n_j}$, for each $j=1,2$. We cluster the whole set of vectors of the two corpora, say $\mathcal{X}:=\mathcal{X}_1 \cup \mathcal{X}_2$, and denote as $\{\mathcal{X}^{k}\}_{k=1}^m$ the $n$ clusters returned by the algorithm. Note that we cope with \emph{hard} clustering methods, i.e.,  $\cup_{k=1}^n \mathcal{X}^{k}=\mathcal{X}$ and $\mathcal{X}^{k_1} \cap \mathcal{X}^{k_2} = \emptyset$ for each $k_1,k_2=1,\ldots,m$, with $k_1\neq k_2$. For each cluster $\mathcal{X}^k$ we count how many of its elements belong to $\mathcal{X}_1$, say $n_{1,k}$, and to $\mathcal{X}_2$, say $n_{2,k}$. We call \emph{impure} a cluster such that both $n_{1,k}>0$ and $n_{2,k}>0$. As we regard the clusters as equivalence classes for the abstract notion of \emph{sense}, if all the $m$ clusters are impure it means that no new senses appeared in $\mathcal{C}_2$ and no new senses have been lost from $\mathcal{C}_1$ to $\mathcal{C}_2$. If this is not the case we might have new senses in $\mathcal{C}_2$, i.e., there is at least a $k$ such that $n_{1,k}=0$, or, vice versa, an old sense has been lost, i.e., there is at least a $k$ such that $n_{2,k}=0$. Following the guidelines of the SemEval shared task, we might set a lower bound $\underline{n}$ to the number of occurrences of a word in a cluster before deciding to regard it as a new sense. If this is the case the above conditions for the counts equal to zero should be replaced by $n_{j,k} < \underline{n}$. Overall, this procedure corresponds to a sound algorithm to address (S1). We refer to it as M1S1.

Regarding (S2), after the clustering, we might define a random variable $S$, to be called the \emph{sense} variable, whose $m$ states are in one-to-one correspondence with the clusters. The variable denotes how likely is finding an occurrence of $w$ with sense $S$ in a corpus. Accordingly, we might use the counts $\{n_{j,k}\}_{k=1}^m$ to learn a probability mass function $P_j(S)$ for each $j=1,2$. Following a Bayesian approach, based on a Laplace uniform prior with equivalent sample size $\sigma>0$ \cite{gelman2013bayesian}, we have:
\begin{equation}\label{eq:bayes}P(s_k) = \frac{n_{j,k}+\frac{\sigma}{m}}{\sum_{k=1}^m n_{j,k}+\sigma}\,.
\end{equation}
In such a probabilistic setup, the semantic shift of the target word $w$ between the two corpora can be therefore described by the dissimilarity between the mass functions $P_1(S)$ and $P_2(S)$. We measure that by the Shannon-Jensen distance $SJ$, i.e., a symmetrization of the popular Kullback-Leibler divergence. This semantic shift of $w$ corresponds therefore to the distance $\delta$, with $\delta:=SJ(P_1,P_2)$ and  $SJ(P_1,P_2):=\frac{1}{2}[KL(P_1,P_2)+KL(P_2,P_1)]$ and $KL(P_1,P_2):=\sum_{k=1}^m P_1(s_k) \ln \frac{P_1(s_k)}{P_2(s_j)}$. Note that with the Bayesian smoothing in Equation (\ref{eq:bayes}), we cannot have zero probabilities and degenerate values in the computation of the distance. The overall procedure gives an algorithm to address subtask S2, as this corresponds to sort the elements of $\mathcal{W}$ with respect to their value $\delta$. We refer to this procedure as M1S2.

\section{Method 2: Separate Clustering of the two Corpora}
In this section, while still focusing on a given target word $w\in\mathcal{W}$, we consider a different approach based on the separate clustering of the two set of vectors $\mathcal{X}_1$ and $\mathcal{X}_2$. Let $\{\mathcal{X}_1^{k_1}\}_{k_1=1}^{m_1}$ and $\{\mathcal{X}_2^{k_2}\}_{k_2=1}^{m_2}$ denote these two sets of clusters. As in the previous method we regard each cluster as a representative model of a sense. Yet, unlike the previous case, here we need to define a map between the clusters of the first corpus and those of the second. As discussed before we adopt the Euclidean distance between the vectors as a proxy indicator of semantic similarity. In order to cope with single numerical values, for the sake of simplicity, we represent each cluster with its center of mass. Let $\tilde{x}_1^{k_j}$ denote the center of mass of $\mathcal{X}_j^{k_j}$ for each $k_j=1,\ldots,m_j$ and $j=1,2$. If $m_1=m_2$, i.e., the two corpora have the same number of clusters, we can reduce the identification of the map between the clusters of the two corpora to a minimum weight matching in a complete bipartite graph, whose nodes are associated with the two sets of clusters and whose weights are the Euclidean distances between the centers of mass. If this is not the case and, for instance, $m_1>m_2$, we add $m_1-m_2$ \emph{dummy} clusters to the second corpus and set to zero the weights for all the arcs connecting these elements. We similarly proceed if $m_2>m_1$.

The optimal matching minimizing the sum of the weights can be computed in cubic time with the classical Hungarian algorithm \cite{kuhn1955hungarian,jonker1987shortest} and the results is a one-to-one correspondence between the clusters, no matter whether proper or dummy, of the two corpora. As a dummy cluster in a corpus has zero distance from all the clusters of the other corpus, the matching returned by the Hungarian algorithm is properly minimizing the distance between the proper clusters. Two proper clusters in the two corpora matched by the algorithm are intended as representative of the same sense. Proper clusters of a corpus pointing to dummy cluster are regarded instead as a new sense appeared in the second corpus only, or old sense occurred in the first corpus only. 

After the matching, we define a single clustering with $m:=\max\{m_1,m_2\}$ clusters and proceed exactly as in the previous section. In practice, the vectors of two clusters matched by the Hungarian algorithm are assigned to a single, impure, cluster, while those linked to dummy clusters produce pure clusters. We term M2S1 and M2S2 the two algorithms corresponding to the approach discussed in this section to address the two subtasks. Next section describes the experimental analysis and evaluation results.

\section{Method 3: An alternative approach for Subtask 2 (S2)}

As an alternative to the previously explained procedure for handling (S2), based on Bayesian approach and Shannon-Jensen divergence, we consider another approach, exploiting only the number of word occurrences per cluster and corpora. More precisely, assuming that we have $\mathcal{K}$ clusters in total and $\forall k \in \{1,.., \mathcal{K}\}$ already calculated $n_{1,k}$ and $n_{2,k}$ from each corpora (regardless whether clusters come from single clustering in M1 or after performing optimal cluster matching in M2), we define the coefficient of semantic change of the word (ranking in (S2) terminology), as:
 $$\frac{1}{2S_1S_2} \sum\limits_{k=1}^{\mathcal{K}}{|S_2\cdot n_{1,k}-S_1 \cdot n_{2,k}|} $$
where $S_1\!=\!\!\sum\limits_{k=1}^{\mathcal{K}} {n_{1,k}}$ and $S_2\!=\!\!\sum\limits_{k=1}^{\mathcal{K}} n_{2,k}$. Let us assume that word $w$ has $p$ occurrences in both corpora. It is trivial to see that in the case with $\mathcal{K}=2$ and clear cut between corpora (e.g. all occurrences in cluster 1 belong to $\mathcal{C}_2$ and all occurrences in cluster 2 belong to $\mathcal{C}_1$, i.e. $n_{2,1}\!=\!n_{1,2}\!=\!p, n_{1,1}\!=\!n_{2,2}\!=\!0$, our coefficient equals 1, which indicates complete change of sense. Likewise, if the distribution of occurrences is uniform ($n_{*,*}\!=\!p/2$), it yields 0, meaning no sense change. We denote these two new procedures for (S2) for M1 and M2 as NM1 and NM2, respectively.

\section{Experimental Analysis}
Experimental analysis is performed according to the rules posed by SemEval2020 challenge\footnote{https://competitions.codalab.org/competitions/20948\#learn\_the\_details-overview} organizers, using provided corpora (2) and baselines (3). Corpora are provided in four languages: English \cite{alatrash2020ccoha}, Latin \cite{mcgillivray2013tools}, German \cite{textarchiv2018grundlage} and Swedish \cite{adesam2019exploring}. Table \ref{Tab:corpora} provides brief statistics for given corpora stating the number of target words (NTW), the total and average number of sentences containing target words (NSTW) per each language. The three baselines provided are: normalized frequency difference (FD), count vectors with column intersection and cosine distance (CNT+CI+CD) and a random baseline always predicting a majority class (RND/MC) - for details see \cite{schlechtweg2019wind}.

For evaluation purposes (as instructed by SemEval guidelines), accuracy is exploited for (S1), while Spearman coefficient, taking values between -1 (corresponding to negative correlation) and 1 (perfect correlation), was used for (S2). More details can be found in the system description paper \cite{schlechtweg2020semeval}. 

It is worth mentioning that the upper bound for the number of clusters for K-means which could be retrieved by the silhouette score was set to 10.

As explained before, for (S1), the idea was to compare the number of elements of each cluster $\mathcal{X}^k$ coming from different corpora, say $n_{1,k}$ and $n_{2,k}$, and claim a change in senses if $\exists k\!: n_{1,k}\!=\!0 \;\vee\; n_{2,k}\!=\!0$. This indeed was the procedure applied for Latin corpora. For other languages (with larger sizes of corpora), following the guidelines of the SemEval shared task, additional restrictions in terms of lower ($\underline{n}$) and upper bounds ($\bar{n}$) were set, with the following purpose: word is considered as gaining a new sense, if $\exists k\!: n_{1,k}\!\le\!\bar{n} \;\wedge\; n_{2,k}\!\ge\!\underline{n}$ (and vice versa for losing a sense). Additionally, suggested values for these bounds were set to $\underline{n}=5$ and $\bar{n}=2$.

The code\footnote{The source code is available at: https://github.com/vanikanjirangat/SST\_BERT-SEMEVAL\_TASK1} is implemented in Python using Scikit \cite{sklearn-api} and Transformers \cite{Wolf2019HuggingFacesTS} library.

\begin{table}[h]
\begin{center}
\scalebox{0.75}{
\begin{tabular}{ccccccccc}
\toprule
\bf \multirow{2}{*}{Language} & \bf \multirow{2}{*}{NTW} & \multicolumn{3}{c}{ $\mathcal{C}_1$} & & \multicolumn{3}{c}{\bf $\mathcal{C}_2$}  \\ \cline{3-5} \cline{7-9}
 & &  \bf total (NSTW) & \bf average (NSTW) & \bf span (years) &&  \bf total (NSTW) & \bf average (NSTW) & \bf span (years)\\ \midrule
English & 37 & 24917 & 673.43 & 1810-1860   && 29088 & 786.16 & 1960-2010   \\
Latin & 40 & 26912 & 672.80 &  -200-0  && 126081 & 3152.03 &  0-2000   \\
German & 48 & 78845 & 1642.60 & 1800-1899 && 68621 & 1429.60 & 1946-1990   \\
Swedish & 31 & 83703 & 2700.10 &  1790-1830   && 241525 & 7791.13 & 1895-1903\\
\bottomrule
\end{tabular}
}
\end{center}
\caption{\label{Tab:corpora} Summary statistics of corpora with respect to target words for different languages}
\end{table}

The experimental results on the corpora over the two subtasks (S1) and (S2) are reported using the methods M1 and M2, in Table \ref{Tab:results}. Results are shown for the four target languages and overall, as well as compared with provided SemEval baselines. Best results per language and subtask are underlined. Overall best results per subtask are denoted in boldface. As can be seen, except for the Latin, the proposed methods are outperforming all baselines on (S1) with the procedure M1S1 being the best for German and Swedish and M2S1 for English. On the other hand, for (S2), results are quite corpus/language dependent, M1S2 scores best for English, M2S2 for Swedish, while baseline 2 (CNT+CI+CD) wins over all the others for Latin and German. Overall, Method 2 outperforms its competitors on both subtasks. The performance with the contextualized embeddings is actually comparable with the baseline approaches in many cases. This could be the fact that pre-trained embeddings from BERT may not be completely suitable for representing meaningful sentence vectors for clustering \cite{reimers2019sentence}. These factors have to be investigated in the future.
%
%
\begin{table}[htp!]
\begin{center}
\scalebox{0.85}{
\begin{tabular}{cccccccccccccccc}
\toprule \bf \multirow{3}{*}{
\begin{tabular}{@{}c@{}}  Language  \end{tabular}
} && 
\multicolumn{2}{c}{\multirow{2}{*}{\bf Method 1}} && 
\multicolumn{2}{c}{\multirow{2}{*}{\bf Method 2 
}} && \multicolumn{8}{c}{\bf SemEval Baselines}  \\ \cmidrule{9-16} 
&&  &  && & && \multicolumn{2}{c}{\bf FD} && \multicolumn{2}{c}{\bf CNT+CI+CD} && \multicolumn{2}{c}{\bf MC} \\ \cmidrule{3-4} \cmidrule{6-7} \cmidrule{9-10} \cmidrule{12-13} \cmidrule{15-16} 
&& \bf (S1) & \bf (S2) && \bf (S1) & \bf (S2) && \bf (S1) & \bf (S2) && \bf (S1) & \bf (S2) && \bf (S1) & \bf (S2) \\ \midrule
English && 0.541 & \underline{0.028} && \underline{0.622} & -0.008 && 0.432 & -0.217 &&	0.595	& 0.022	&& 0.568 & / \\
Latin && 0.375 & 0.253 && 0.625 & 0.253 && \underline{0.650} & 0.020 && 0.525 & \underline{0.359} && 0.350 & / \\
German && \underline{0.708} & 0.176 && 0.625 & 0.176 && 0.417 & 0.014 && 0.688 & \underline{0.216} && 0.646 & / \\
Swedish && \underline{0.742} & 0.275 && 0.677 & \underline{0.321} && 0.258	& -0.15	&& 0.645 &	-0.022	&& \underline{0.742} & / \\ \midrule
Overall && 0.591 & 0.183 && \bf{0.637} & \bf{0.185} && 0.439 & -0.083 && 0.613 & 0.144 && 0.576 & / \\
\bottomrule
\end{tabular}
}
\end{center}
\caption{\label{Tab:results} Performance of the proposed methods and different SemEval baselines. The baselines correspond to normalized frequency difference (FD), count vectors with column intersection and cosine distance (CNT+CI+CD) and majority class(MC)} 
\end{table}


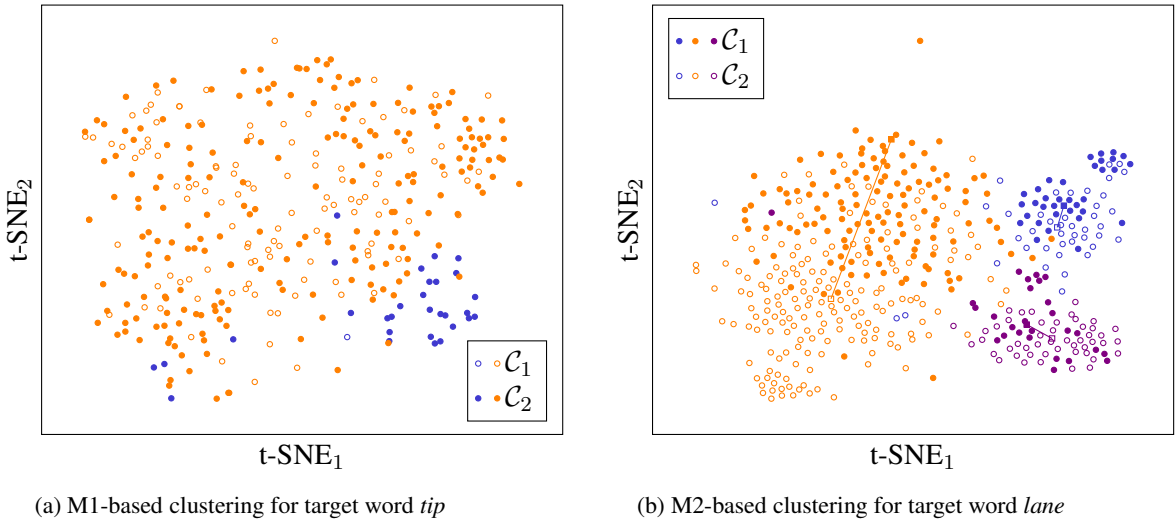
\begin{figure}[h]
\begin{subfigure}[t]{0.4\textwidth}
\begin{tikzpicture}
\definecolor{myblue}{rgb}{0.247,0.224,0.812}
\definecolor{myorange}{rgb}{1.000,0.500,0.000}
\begin{axis}[
legend pos=south east,
legend columns=2, 
legend style={/tikz/column 2/.style={column sep=1pt,},},
xtick=\empty, ytick=\empty,
xlabel=t-SNE$_1$,
ylabel=t-SNE$_2$]
\addplot[only marks,mark=o,myblue,mark size=1pt] coordinates {(5.2870936,-13.249211)};
\addlegendentry{};
\addplot[only marks,mark=o,myorange,mark size=1pt] coordinates {(1.4550648,11.576058)(6.7977514,10.786149)(2.2756948,-1.4116857)(1.6548021,-4.841949)(3.961691,1.464099)(-11.22501,-7.1336713)(4.7085204,-8.536829)(-3.8579032,-18.241827)(-9.22686,10.634559)(-11.571849,3.520439)(2.1419973,5.8517156)(-10.547626,2.145415)(7.3158474,-3.6926904)(-3.8491888,-8.525413)(5.6793118,2.100787)(0.14455013,-8.64062)(-4.224431,-4.9528084)(3.2651982,2.3033628)(-0.18577251,8.422214)(-2.3646655,-14.043809)(-11.311605,12.43909)(-14.429762,7.962687)(-2.3533475,-4.3018985)(-1.0029757,10.462579)(-18.241377,-10.76623)(-4.505302,-2.8429751)(-10.357799,3.4391518)(8.126097,3.7838233)(4.154503,-7.1130824)(-7.663965,5.6053786)(-2.3043225,-10.546591)(8.866294,11.240556)(0.33396342,-16.665636)(11.389733,-0.3423668)(-14.139751,12.618565)(-9.508172,-8.983612)(-7.156751,1.748699)(-1.414841,-3.1025126)(-4.3181705,10.300499)(-17.41803,-2.9624748)(-1.550434,-1.1646628)(-9.674034,3.7724411)(-7.705224,5.7644663)(14.900568,4.829709)(-0.6191451,4.693395)(16.281284,0.33396176)(-13.456268,13.978088)(-1.9640329,-2.561689)(14.026667,13.7140665)(19.228464,13.635562)(-0.20634899,0.5761092)(-5.7422247,5.9152265)(4.3470397,12.392783)(-4.581869,-3.230993)(10.937451,12.00637)(-10.344611,-6.8261113)(-1.879596,19.72717)(11.590052,7.5369716)(14.622437,7.992425)(-15.695539,6.692012)(9.419533,6.980601)(-11.3117,12.439279)(-15.960079,11.438719)(10.372122,-4.550843)(-16.954216,7.930347)(4.799324,4.2857566)(-4.248809,9.119898)(5.9900713,0.66071564)(-4.7458043,-7.8899736)(-14.752851,-9.937006)(9.067106,1.8232354)(-6.091263,-8.34522)(-15.023131,7.400286)(18.304102,5.1554413)(17.173388,4.1705294)(-12.215918,-6.409133)(-1.5700743,-10.599841)(-20.626385,9.058568)(-5.507336,2.6343253)(-3.9711432,3.3030322)(7.3671813,-4.1339984)(8.411926,4.9377074)(7.6924906,-2.8429804)(8.415514,0.3854158)(-8.516613,-5.7375965)(-10.1442795,-13.248971)(-13.190474,-10.049439)(6.480737,1.7971262)(-2.5181105,-3.539729)(-4.439475,-7.3132)(-13.446104,6.9316034)(-13.57022,10.661177)(-8.6438675,14.264063)(-14.588385,11.953835)(-18.016918,8.2590475)(4.6998076,5.5041304)(-3.8383918,-4.783262)(-6.0783753,-0.90867543)(1.0269775,-6.868699)(7.9587026,15.38383)(-4.518632,3.830029)(7.0609336,2.3362134)(-19.852144,8.968295)(13.916752,6.3143787)(-5.9162245,4.9447613)(7.9539886,-8.341967)(-13.451484,-7.589535)(-14.442808,8.730771)(2.1941965,7.189955)(-8.761556,2.9958303)(-7.0851464,-15.95051)(3.0535567,1.0521002)};
\addlegendentry{$\mathcal{C}_1$};
\addplot[only marks,mark=*,myblue,mark size=1pt] coordinates {(-13.83632,-16.628471)(-12.112781,-20.065868)(13.957378,-13.958332)(12.807544,-8.558275)(4.2929816,0.28038004)(17.328949,-11.151867)(16.2896,-4.4757996)(9.681873,-10.983989)(-5.986152,-13.501397)(17.156775,-10.558146)(5.653686,-10.564513)(14.992565,-10.838266)(-12.795801,-16.247047)(13.55688,-13.215272)(14.519073,-10.684855)(17.939415,-8.786013)(17.904255,-10.98283)(13.459014,-10.202836)(14.092268,-7.0968537)(13.4203415,-7.091887)(14.505163,-13.7262745)(13.941095,-4.2473493)(16.079361,-6.1257267)(12.620461,-5.6191254)(10.537869,-9.805575)(9.481263,-12.694698)(4.0572777,-4.9431677)(9.268928,-11.155895)(15.74731,-12.088061)(9.450204,-13.699842)};
\addlegendentry{};
\addplot[only marks,mark=*,myorange,mark size=1pt] coordinates {(11.277067,-4.5619955)(-8.587279,-15.122726)(-7.4634876,-8.164782)(-3.4341693,14.178581)(-17.614336,6.917777)(17.352955,9.9462385)(-2.5427785,14.412125)(-8.81499,-3.6380024)(9.892843,-8.596448)(16.988552,11.023581)(-10.1925535,5.9892845)(17.44277,5.6296053)(14.532047,2.5360377)(5.716479,15.00866)(2.1799555,3.511904)(-4.961027,16.353472)(-12.629746,-8.937966)(8.598187,6.852765)(-14.331964,-4.062449)(1.1389478,-2.086205)(16.392023,6.641684)(20.773756,8.810814)(16.428074,8.838593)(-17.336811,-5.755054)(20.348202,6.737033)(-18.44615,-10.331293)(-5.019148,3.9854574)(2.843349,-11.085655)(-14.423784,-5.2740135)(10.428391,-7.0556073)(-10.426346,4.4645576)(-4.3454256,0.41539186)(-17.438303,-12.984311)(3.3428686,-1.9469892)(-18.754652,6.3381925)(-19.38787,-11.66222)(3.3697538,8.966208)(3.6613877,17.658722)(18.687504,2.8223479)(7.8163853,-9.077399)(18.233929,3.5342538)(-3.2566254,-5.656799)(-4.6226597,8.571191)(16.994566,7.880353)(1.8989778,-8.524785)(4.4558806,15.701227)(4.1816134,9.356098)(9.357564,-13.9153595)(-6.2545547,-5.1855044)(-16.022099,4.8721585)(-16.783585,-6.192)(4.924705,11.053643)(3.2637491,17.115667)(-1.0965021,-7.9883146)(-16.495428,-9.61968)(15.342496,12.807974)(-1.9888804,14.934894)(9.71493,0.66960067)(11.787,11.856967)(-7.7450056,-10.328239)(20.677908,6.515455)(-20.232784,-0.18118986)(-20.236801,-0.18445311)(-13.470426,-2.5587797)(11.418271,9.403467)(10.24747,9.551726)(-11.202246,-8.00354)(1.2694016,16.59612)(22.310043,3.800859)(-13.612984,3.3771632)(5.839016,-3.2920332)(4.016079,14.504386)(20.609827,10.870333)(6.092451,-4.713932)(-3.2825878,16.526743)(-11.074119,-15.28298)(4.9379754,9.111787)(-0.29340512,13.691092)(11.453341,1.892598)(-6.3869176,-19.4566)(-18.755974,10.448723)(7.6179333,-7.415759)(-13.580809,-12.095375)(-15.831795,8.707836)(-16.6639,-11.668378)(-16.681862,9.702992)(-11.347959,-11.724057)(-1.5322056,6.690621)(6.5025454,8.044973)(-15.576553,-11.173355)(-20.646648,9.61184)(-12.4837265,-4.4505825)(14.69961,5.7199397)(17.66304,6.489138)(-3.2455192,8.4441595)(10.377936,3.8559015)(-9.824628,-3.3715773)(19.301626,6.458098)(4.3458066,-16.920282)(14.1849,10.224127)(5.727932,-6.8938527)(-7.1365333,-9.401938)(13.618091,12.413525)(16.221096,0.4511297)(-17.75367,2.3449473)(11.806754,4.0107408)(-12.400886,-13.9077215)(-8.331995,4.539517)(6.7706175,5.868031)(13.33683,0.55337656)(10.784448,-3.3175087)(11.86898,4.8787155)(14.359351,12.452881)(6.2444487,-1.4802421)(16.61669,5.169594)(-19.467688,10.88446)(-2.4001117,12.762579)(-8.878332,-13.6107435)(7.881121,12.707183)(10.317356,12.822962)(-7.315634,-11.27299)(12.192669,-1.7123842)(-14.351043,-1.5808133)(5.839841,4.103186)(-8.023787,7.2878213)(-14.961851,-2.0655525)(-12.942583,10.999631)(-12.222925,-13.403754)(-12.061107,-2.712332)(9.968291,0.67958087)(13.911407,11.712156)(18.725199,8.152598)(-12.730186,-8.751026)(0.3228636,16.729525)(-6.2123823,-5.199758)(-3.8108618,0.9919987)(-9.227063,-2.3243663)(-15.15964,9.9764)(-10.326542,7.8205295)(0.56193924,16.65871)(-13.609616,14.756366)(2.6010501,15.3640995)(18.410706,8.836906)(-7.622465,-20.104467)(8.858055,-7.937307)(11.458799,14.515512)(19.68855,4.58703)(-12.526336,8.130004)(-8.13419,14.323501)(-0.68147147,-4.5826287)(-11.90079,15.292106)(14.3487,-3.0502644)(-3.35634,-1.5095677)(-4.9143157,1.3611454)(-15.351239,1.1543341)(-15.720598,-6.4738116)(10.030722,-1.5891035)(-11.771024,-14.264868)(7.5692196,9.662696)(-10.480212,-0.8326724)(-10.31592,-17.043043)(8.0678215,9.214266)(6.0532203,8.6483135)(11.305859,4.6798563)(8.232645,-1.6293901)(-10.429463,10.2574625)(17.844023,7.6435103)(-12.790446,-11.98395)(6.326667,15.348258)(11.206078,-6.1571884)(-15.248067,-13.581069)(-12.19199,-18.039639)(-10.461351,-3.4091055)(-5.378335,13.613514)(-12.914876,1.1454095)(0.5482563,-2.909911)(9.649908,-3.2219512)(7.749536,-5.4961696)(9.126841,-5.090521)(17.621998,8.9948845)(-7.8390937,-9.77419)(-10.850987,-10.208579)(0.700841,-13.467621)(-6.464289,-19.456566)(15.683557,14.028204)(-8.575973,0.37163717)(-18.852798,-7.3615804)(-6.4767637,-10.238482)(-7.3016043,-2.0447662)(16.37646,-6.5225134)(-13.189626,4.5799594)(-6.655775,-11.998862)(-16.59886,13.499295)(-12.535525,-10.558451)(19.23342,8.991463)(-14.8135,-8.776763)(5.286761,12.170262)(10.97468,8.590159)(5.881335,12.652862)(-9.0298605,-12.880015)(-9.163084,-10.807677)(1.9845659,12.792662)(-15.381305,2.1287053)(-12.318115,-18.651928)(2.324318,17.575274)(-16.657967,-12.635897)(3.3867047,-2.6926906)(-1.408421,-7.2234626)(-9.651781,-10.693098)(-9.078723,10.38606)(18.600967,10.970758)};
\addlegendentry{$\mathcal{C}_2$};
\end{axis}
\end{tikzpicture}
\caption{M1-based clustering for target word \emph{tip}\label{fig:m1}}
\end{subfigure}
\quad \quad \quad \quad
\begin{subfigure}[t]{0.4\textwidth}
\begin{tikzpicture}
\definecolor{myblue}{rgb}{0.247,0.224,0.812}
\definecolor{myorange}{rgb}{1.000,0.500,0.000}
\definecolor{mypurple}{rgb}{0.50,0.0,0.50}
\begin{axis}[
legend pos=north west,
legend columns=3, 
legend style={/tikz/column 3/.style={column sep=1pt,},},
xtick=\empty, ytick=\empty,
xlabel=t-SNE$_1$, ylabel=t-SNE$_2$]
\addplot[only marks,myblue,mark size=1pt,mark=*] coordinates {(117.477234,120.21246)(108.10185,67.40573)(89.50146,94.71144)(70.95865,68.18676)(89.069275,39.159073)(87.12023,74.86762)(103.67522,61.15948)(81.20274,29.149828)(98.86626,92.59337)(139.77121,121.15515)(125.49593,106.114555)(124.616776,123.95342)(96.34083,73.61686)(85.60708,65.46434)(133.7084,106.16026)(138.771,47.773514)(103.14039,72.9691)(92.86551,66.51694)(109.092224,19.272701)(69.92084,47.65017)(120.38348,110.44199)(82.47485,47.697144)(132.94145,125.298904)(109.69397,52.92987)(112.2926,60.369404)(144.30011,112.338554)(72.46309,79.51622)(132.53923,118.05409)(124.7478,116.08242)(94.206474,57.27642)(81.98386,84.76999)(112.81292,73.68358)(80.14878,69.636795)(74.67607,55.595303)};
\addlegendentry{};
\addplot[only marks,myorange,mark size=1pt,mark=*] coordinates {(-22.60366,48.12744)(14.452265,94.57838)(-50.609688,52.05169)(-116.268936,61.96892)(47.494938,79.89766)(-9.735723,121.91609)(-22.963373,132.28279)(32.91563,70.653595)(11.24025,107.31081)(-64.900116,123.56188)(-96.81834,11.543009)(28.677534,-12.964641)(-40.044643,46.496643)(-44.232216,5.0956144)(7.9639993,-2.8917859)(78.2026,9.29638)(5.112629,60.58357)(49.112225,50.206726)(45.179348,99.66392)(-80.81874,37.714897)(-33.978878,97.05455)(-18.088102,77.27744)(-41.37874,149.28152)(-116.17363,50.579994)(-20.00138,110.36648)(11.0203705,-122.323616)(-33.011974,20.884922)(-2.1108274,110.52695)(-50.227886,63.28494)(20.10013,100.11841)(-58.337975,4.271691)(-21.572079,98.82971)(-49.09069,-98.76635)(1.3994642,92.84687)(-27.61368,108.02165)(-1.4986933,18.523777)(-10.20728,-17.903002)(-36.99518,111.34929)(-51.371216,11.049213)(91.08873,30.36173)(-80.87748,82.68777)(-33.825424,-9.901856)(-22.415613,-7.112704)(-112.22925,66.86507)(-8.283165,66.93306)(-12.261737,56.368286)(-9.352864,35.597523)(-85.985565,42.12225)(-53.70352,87.97275)(-51.35651,70.335045)(8.225936,89.790146)(20.885859,-7.302259)(-63.283607,53.911343)(-44.389656,-9.374246)(-45.58471,39.246723)(63.766476,13.764656)(-36.95602,129.51625)(-11.258979,46.454193)(-60.42845,38.87359)(-78.93311,117.13866)(-3.2564187,135.67844)(-18.206507,9.771501)(-2.077838,11.303073)(54.18333,42.46573)(-29.170738,118.72475)(-31.673258,14.766068)(13.386139,51.918858)(-37.582993,124.01653)(-108.39557,5.277584)(33.011276,82.765076)(-69.20582,46.846264)(11.123904,31.437193)(-13.585397,26.873814)(11.690638,45.231693)(-86.99801,70.33228)(24.662766,27.447168)(-92.72252,84.83784)(-70.32681,90.16786)(6.0058713,0.23231968)(-67.27929,63.477127)(-26.109146,54.527073)(-48.155254,21.259058)(-5.4526877,85.72871)(-19.64114,59.531845)(-32.498955,47.81196)(-13.237247,91.039856)(-26.775927,82.483894)(56.860703,26.934898)(-46.805534,94.97493)(-25.607689,2.807653)(5.1570334,124.33054)(0.16896164,27.149044)(-41.502876,103.10654)(-56.509636,103.844475)(-110.03464,99.315796)(-62.889812,-30.241907)(-24.160856,93.07746)(12.077714,75.59168)(-114.05607,41.22862)(-74.76704,46.00325)(15.213035,137.31331)(-8.470373,103.31265)(-50.38864,-1.7983865)(-31.282848,138.98329)(-22.546703,115.19768)(-101.68994,42.394085)(-9.23216,78.559555)(38.979195,96.89876)(-92.67368,47.371426)(11.119191,13.182978)(20.879213,85.36094)(-30.431969,38.995937)(-24.148958,29.548258)(-103.79649,80.15151)(-7.221492,-13.520582)(-61.554676,78.9565)(22.248362,-15.753305)(-52.07416,121.830055)(-88.13419,80.12614)(4.8055663,-19.238401)(-72.74829,78.89635)(-14.030441,144.52449)(-34.05883,58.242065)(-85.15562,-22.83198)(-21.527626,126.573586)(35.042187,47.363846)(2.2124805,247.59682)(-40.764526,67.447334)(-49.055416,28.854654)(-53.40685,-28.614197)(-41.505173,79.22587)(23.39721,-25.120964)};
\addlegendentry{};
\addplot[only marks,mypurple,mark size=1pt,mark=*] coordinates {(128.71066,-103.52959)(69.03289,-5.7391167)(-98.208755,59.09273)(73.14956,-60.100002)(52.23063,-54.995586)(75.19494,-14.006622)(102.86882,-103.73408)(126.86505,-95.91748)(64.59657,-66.62815)(58.993126,-75.93168)(92.7213,-113.48588)(121.08742,-91.435425)(82.7902,-15.618419)(86.96367,-11.523749)(108.51807,-73.27561)(37.8818,-40.264267)(39.610825,-45.272945)(80.87398,-9.207805)(77.78321,-70.61811)(88.751656,-39.33811)(76.7917,-23.868668)(54.395187,-81.05526)(76.22001,-78.0303)(101.99165,-70.37647)(94.45753,-101.30147)(69.23616,-72.54691)(91.71575,-51.212532)};
\addlegendentry{ $\mathcal{C}_1$};
\addplot[only marks,myblue,mark size=1pt,mark=o] coordinates {(46.81067,66.00405)(123.301834,33.97617)(-7.809461,-53.545135)(78.26667,38.77603)(83.79618,56.790657)(100.23588,53.402344)(87.60469,20.01483)(113.755646,40.409344)(130.0408,97.09696)(98.590744,24.599485)(97.63264,35.97974)(137.16095,112.51685)(132.15703,60.468285)(104.892296,44.30316)(61.996403,57.75766)(96.32606,12.494298)(124.315544,44.0375)(121.5347,66.506584)(56.293247,94.23469)(115.323944,83.0914)(130.1742,111.94119)(119.87111,55.534737)(98.20385,-27.38052)(140.30916,103.79167)(97.46361,45.40233)(126.16714,77.94941)(113.27201,31.76588)(106.68156,34.388985)(104.02235,80.93565)(74.18018,20.399025)(90.24862,49.258915)(104.960266,27.302803)(68.77648,31.03323)(54.249218,3.510997)(87.12273,-1.1058397)(94.1728,81.22295)(136.35655,29.51929)(-13.656565,-56.838623)(-136.70596,69.951836)(100.03641,-8.709117)(119.07695,21.632153)(115.48388,12.4832325)};
\addlegendentry{};
\addplot[only marks,myorange,mark size=1pt,mark=o] coordinates {(-104.70656,-130.58757)(-16.589262,3.2674317)(37.62511,118.1343)(-76.76404,-17.362637)(-82.44122,-35.726715)(-99.5,-10.647646)(-98.99731,-71.13505)(-47.220535,-45.632282)(-76.55595,-27.335451)(-88.38944,-94.418434)(-49.1144,111.73359)(-35.320797,-72.69971)(-45.262432,-28.959818)(-4.2064724,-36.96114)(-17.83411,-84.87852)(-120.362854,-5.1738)(-73.012856,101.11806)(-28.463572,-78.58441)(-79.533554,-135.00441)(-111.6549,19.296959)(-88.25895,-123.34187)(-32.703526,87.6455)(-80.65379,-45.649105)(0.44513693,42.52232)(-89.176155,-7.8182187)(-101.11322,-48.381058)(-27.275333,-28.929544)(-81.117256,-8.441413)(-55.98704,-50.202316)(-7.489248,-111.6467)(-83.719444,-2.1003852)(-87.66352,-71.894135)(-91.49563,24.472813)(-94.20001,-33.017586)(-69.985954,27.831903)(-118.84905,-27.336687)(-136.81499,-24.194408)(3.3555512,82.436356)(-66.0795,-16.191757)(-35.66732,-81.28489)(34.281467,-74.804726)(-64.43623,-39.169575)(-73.730804,8.135488)(-63.119026,-6.117888)(-70.703476,-5.6571946)(17.6763,66.45603)(-149.36028,1.729231)(-17.278048,-74.368004)(-101.0939,-38.114426)(-33.73118,29.204622)(-93.48599,-41.794453)(41.263668,21.5467)(-2.1932905,53.71364)(-95.039314,-0.9240652)(-12.84008,-88.79514)(-89.69315,100.561646)(-47.249584,-19.451117)(36.66345,34.70064)(-57.398438,21.803196)(-91.71236,-15.97598)(-70.76002,-78.92871)(-78.68888,-80.577805)(-23.1032,-48.8237)(-21.06007,67.78412)(-52.251114,-68.94717)(-42.531578,-54.985325)(-39.306934,-65.11829)(-37.52454,-50.56888)(-93.00852,-101.00441)(-83.615685,-141.44293)(-90.51313,-49.977417)(-44.15767,-80.6059)(-76.52505,-130.42514)(-57.70885,-43.546513)(-74.12091,-70.31671)(-0.09109668,-34.64738)(-41.4287,-38.072765)(-76.22864,-55.79298)(-104.04692,-138.80847)(-56.391644,-14.192353)(-123.93784,-63.252037)(-3.06054,-80.091644)(-3.9194698,-73.876686)(-72.258934,-113.56186)(-5.2886367,4.0915184)(-122.81671,31.282213)(-108.741295,-55.190742)(-99.44112,-114.70514)(-96.528694,-128.10347)(12.283918,20.860334)(-56.08931,-143.57498)(-112.8273,-30.292171)(8.075631,-75.466034)(-131.63402,-60.139275)(-65.998405,-52.398903)(-30.14533,-95.11158)(-149.14395,-4.847346)(-78.61988,55.7263)(-64.827965,-71.19841)(-52.220264,-59.28457)(-90.82597,-61.51381)(-97.61522,-136.16714)(-14.077097,-4.4045525)(-80.22568,-88.21648)(-68.33599,-25.669031)(60.510967,-26.406342)(-101.081,-122.29679)(-94.00839,-118.757774)(-19.75653,37.528103)(27.033691,58.3004)(0.39764312,72.56321)(-64.42407,-96.26174)(-96.03122,-55.101055)(3.9330919,-41.29269)(49.937126,-13.91506)(56.86523,-41.281593)(-113.235344,-105.899536)(22.964428,39.410145)(34.71279,10.299372)(-81.83097,-107.1826)(-45.240494,58.811928)(14.2312765,-49.536076)(-20.249197,19.547958)(22.887999,7.1786356)(-80.573814,21.444643)(-102.652565,31.535055)(25.351261,17.509935)(-59.68284,-23.2279)(-91.41756,-134.35387)(-43.075428,24.576448)(-77.85869,2.6304247)(-101.90094,-24.644886)(-62.84208,-62.74762)(-72.45691,-139.30383)(-105.077774,-20.090826)(-73.78357,-94.919846)(-88.932434,-27.993551)(-108.00298,-69.42754)(-79.62772,72.12692)(-67.17464,-134.80272)(3.8236434,-60.778915)(-83.79826,12.369517)(-37.583565,-26.274467)(-27.221745,-44.26447)(-51.53601,-36.839775)(-86.1662,-56.86932)(-98.7704,-94.329056)(-91.2449,-141.68098)(-83.13395,-65.07985)(-129.2869,3.2702236)(-27.5096,-69.43155)(-111.92778,-45.797028)(-108.115234,-122.69254)(-102.45304,-60.923874)(-17.43057,-40.570446)(-9.768287,-35.61496)(-32.9641,-107.37417)(-60.412453,-81.83631)(-72.41845,-35.03546)(-98.36893,-145.06796)(-44.040104,87.99009)(-29.995945,70.947235)(-34.641727,3.9724293)(-114.4985,-12.460297)(-39.603283,12.402886)(-86.22765,-131.95265)(-32.99831,-62.01486)(-23.930885,-23.957766)(-73.59813,-42.542416)(-113.57636,-65.94614)(-61.53125,14.04951)};
\addlegendentry{};
\addplot[only marks,mark=o,mypurple,mark size=1pt] coordinates {(55.694096,-102.83714)(62.454426,-48.26535)(68.32483,-84.893906)(62.930157,-90.24081)(107.35789,-57.703938)(113.56422,-88.68963)(118.94226,-67.76613)(106.261765,-86.53942)(94.615944,-61.43247)(105.05475,-112.159164)(83.773476,-105.018974)(92.0832,-70.104645)(126.04607,-83.995514)(67.59437,-97.91262)(119.445984,-101.92535)(44.178215,-80.427315)(84.34026,-61.48412)(107.43493,-93.72756)(119.55319,-60.04996)(72.738365,-105.38905)(86.99924,-96.21534)(104.661354,-52.69164)(101.57148,-81.09609)(62.093853,-58.784122)(74.001854,-89.34511)(117.028465,-111.03587)(83.494644,-75.6108)(77.740105,-49.659657)(24.774166,-65.24754)(118.79183,-82.73391)(135.80807,-82.26766)(124.72681,-75.65898)(95.29722,-90.528496)(107.217224,-39.014507)(112.28958,-66.13456)(71.04453,-50.342667)(110.6018,-81.06969)(36.98566,-85.36327)(81.81099,-85.49157)(54.081444,-65.78905)(114.830315,-95.01529)(136.01468,-95.43639)(134.29669,-68.470055)(76.69115,-96.908424)(52.011288,-96.05414)(116.4803,-75.308655)(90.719795,-82.68689)(46.410625,-70.828735)};
\addlegendentry{ $\mathcal{C}_2$}
\addplot[only marks,myblue,mark size=1pt,mark=square*,forget plot] coordinates {(99.42647,66.80688)};
\addplot[only marks,myorange,mark size=1pt,mark=square*,forget plot] coordinates {(-17.187979,139.46199)};
\addplot[only marks,mypurple,mark size=1pt,mark=square*,forget plot] coordinates {(74.2621,-63.702682)};
\addplot[only marks,myblue,mark size=1pt,mark=square,forget plot] coordinates {(94.92448,42.907845)};
\addplot[only marks,myorange,mark size=1pt,mark=square,forget plot] coordinates {(-58.067024,-35.487804)};
\addplot[only marks,mypurple,mark size=1pt,mark=square,forget plot] coordinates {(91.45782,-78.95802)};
 \addplot[mark=none,myblue,forget plot] coordinates { ((99.42647,66.80688) (94.92448,42.907845) };
 \addplot[mark=none,myorange,forget plot] coordinates { ((-17.187979,139.46199) (-58.067024,-35.487804) };
 \addplot[mark=none,mypurple,forget plot] coordinates { ((74.2621,-63.702682) (91.45782,-78.95802) };
\end{axis}
\end{tikzpicture}
\caption{M2-based clustering for target word \emph{lane}\label{fig:m2}}
\end{subfigure}
\caption{Obtained clusterings for target words}
\end{figure}

Figure \ref{fig:m1} shows an example of the application of method M1 for the English target word \emph{tip}, based on 2D t-SNE \cite{maaten2008visualizing} projections. The method produces two clusters, each denoted with a different color. As it can be seen, one of the clusters (orange, $k=1$) is remarkably larger than the other (blue, $k=2$). Additionally, it is also quite impure containing word occurrences from both corpora (more precisely, $n_{1,1}=112$ and $n_{2,1}=211$), while the other cluster contains only 31 instance whose distribution is  $n_{1,2}=1$ and $n_{2,2}=30$. Given that $n_{1,2}=1 \le 2=\bar{n}$ and $n_{2,2}=30 \ge 5=\underline{n}$, method M1S1 correctly detects the sense change for the word \emph{tip}.

Figure \ref{fig:m2} shows an example of the application of method M2 for the English target word \emph{lane} (2D t-SNE projections). The clustering algorithms produce three clusters for each corpus, each denoted with a different color, and the matching algorithm detect the correspondence between clusters minimizing the distances between the centers of mass, depicted as squares in the figure. Matching clusters of the two corpora are depicted with the same color.

The results of the alternative method M3 for subtask (S2) with respect to M1 and M2 and two baselines (FD and CNT+CI+CD), are provided in Table \ref{Tab:new_formula_results}. We can see that NM1 improves results on English and German languages, and overall.

\begin{table}[h]
\begin{center}
\scalebox{0.85}{
\begin{tabular}{ccccccc}
\toprule 
& \multirow{2}{*}{\bf M1} & \multirow{2}{*}{\bf M2} & \multicolumn{2}{c}{\bf SemEval Baselines} & \multirow{2}{*}{\bf NM1} & \multirow{2}{*}{\bf NM2} \\ \cmidrule{4-5}
& \bf  & \bf  & \bf FD & \bf CNT+CI+CD & \bf  & \bf  \\ \midrule
English & 0.028 & -0.008 & -0.217 & 0.022 & \underline{0.159} & 0.037 \\
Latin & 0.253 & 0.253 & 0.020 & \underline{0.359} & 0.231 & 0.333 \\
German & 0.176 & 0.176 & 0.014 & 0.216 & \underline{0.525} & 0.062 \\
Swedish & 0.275 & \underline{0.321} & -0.15	& -0.022	 & 0.141 & 0.095 \\ \midrule
Overall & 0.183 & 0.185 & -0.083 & 0.144 & \bf{0.264} & 0.132  \\
\bottomrule
\end{tabular}
}
\end{center}
\caption{\label{Tab:new_formula_results} Performance of the semantic change coefficient methods NM1 \& NM2 for (S2)} 
\end{table}


\section{Related Work}

The task of identifying words whose meaning has changed over time is well-known and the related literature is, therefore, resourceful with many recent advancements. Albeit, there are still quite some issues to be resolved, primary regarding the methodology and the respective ground truth (missing semantic change annotations). 

As for the latter, the first step is to decide whether to aim for a binary response (equivalent of SemEval2020 subtask 1) or to provide graded ratings of a sense change (equivalent of SemEval2020 subtask 2). Given that in both cases, but particularly with graded rating, inter-annotator agreement rates vary greatly, as evidenced in \cite{erk2009investigations}, establishing a definition of a standard test set is extremely difficult. In \cite{schlechtweg2018diachronic}
a unifying evaluation framework for unsupervised lexical semantic change detection was proposed based on changes in relatedness of word use pairs in each time period. 

Regarding the former, many different approaches for unsupervised lexical semantic change detection have been suggested. A detailed survey of studies can be found in \cite{kutuzov2018diachronic,tahmasebi2018survey}. Most notably, several works \cite{baroni2014don,kim2014temporal,hamilton2016diachronic} showed the benefits of using dense word representations for semantic shift detection. Furthermore, \cite{kulkarni2015statistically} showcased that these outperform the frequency-based methods. However, unlike our work, none of these works exploits clustering. 
More close to our approach, that is, considering clusters as representative semantic areas, are the works of \cite{mitra2014} and \cite{dubossarsky2015bottom}. The main differences are however, that in \cite{mitra2014}, clustering is performed on the level of the ego-network of each word, where the network is constructed based on word co-occurences, while we perform clustering of the word embeddings itself. Additionally, in contrast to \cite{dubossarsky2015bottom} where the authors consider incremental learning of word embeddings in yearly chunks and vary the number of clusters from 500 to 5000, we use silhouette scores to determine the optimal number of clusters per each target word. 

\section{Conclusion and Future Work}

A word can have a different meaning (sense) in different contexts and/or different time periods. Despite being quite extensively studied, the problem of identifying words that have changed their meaning over time, particularly in an unsupervised way, still challenges researchers. 

In this work, we propose two approaches, both of which combine contextualized word embeddings (obtained by BERT) and clustering, differing thus only in the way the clustering has been performed. Considering obtained clusters as proxies for word meanings allows us to quantify the level of change per each target word in four target languages. The obtained results, especially looking overall, across all target languages, where we are outperforming all provided baselines, demonstrate the usefulness of the suggested approach.

As potential directions for future work we plan to investigate various strategies, including different clustering methods and  time-wise comparison of target words nearest-neighbours, in an attempt to identify actual word senses more accurately. Additionally, we would like to further scrutinize how the linguistic particularities of different corpora might have contributed to the variability of the results.



\bibliographystyle{coling}
\bibliography{semeval2020}

\end{document}